%% file: main.tex
\def\year{2019}\relax
\documentclass[letterpaper]{article} %DO NOT CHANGE THIS
\usepackage{aaai19}  %Required
\usepackage{times}  %Required
\usepackage{helvet}  %Required
\usepackage{courier}  %Required
\usepackage{url}  %Required
\usepackage{graphicx}  %Required
\usepackage{scalerel}

\usepackage{mathtools}    
 \usepackage{pifont}% http://ctan.org/pkg/pifont

 \usepackage{fancybox}

\usepackage{booktabs}
\usepackage{color}
\usepackage{amsfonts}
\usepackage{amsmath}
\usepackage{proof}
\usepackage[amsmath,thmmarks]{ntheorem}
\usepackage{relsize}
\usepackage{mathtools}    
\usepackage{paralist}    
\usepackage{pifont}% http://ctan.org/pkg/pifont
\usepackage{mdframed}

\input{commands.tex}
\usepackage{fancyhdr}

\usepackage{graphicx} % Allows including images
\usepackage{booktabs} % Allows the use of \toprule, \midrule and \bottomrule in tables
\usepackage{amsmath} % Allows the use of \toprule, \midrule and \bottomrule in tables
\usepackage{amsfonts}
\usepackage{amssymb}
 \newcommand{\VE}{\ensuremath{\mathcal{V}}}
\newcommand{\Vf}{\ensuremath{\mathcal{V}^f_z}}
\newcommand{\Vz}{\ensuremath{\mathcal{V}_z}}

\newcommand{\CE}{\ensuremath{\mathcal{C}}}
\newcommand{\DE}{\ensuremath{\mathcal{D}}}
\newcommand{\DCE}{\ensuremath{\mathcal{DC}}}
\newcommand{\citelow}[1]{\cite{#1}}
\begin{document}

%%\title{Toward the Engineering of Virtuous Robots}
\title{Toward the Engineering of Virtuous Machines}
%
%% NAVEEN: Do you think we should go with the second of the titles
%% above, because it's more general?  Leave it up to you.  Thx.  //S

%\author{Primary Disciplines: Artificial Intelligence and Philosophy}

\author{Naveen Sundar Govindarajulu $\bullet$ Selmer Bringsjord
  $\bullet$ Rikhiya Ghosh\\
Rensselaer AI \textit{\&} Reasoning Lab, Rensselaer Polytechnic Institue, (RPI)\\
 Troy, New York 12180, USA\\
 www.rpi.edu}
%% NAVEEN: Do we include author names?  If so, we can get away with
%% bullets here (which I did)?  //S
%
% \address{}

\maketitle
\begin{abstract}
\noindent
While various traditions under the `virtue ethics' umbrella have been
studied extensively and advocated by ethicists, it has not been clear
that there exists a version of virtue ethics rigorous enough to be a
target for machine ethics (which we take to include the engineering of
an ethical sensibility in a machine or robot itself, not only the
study of ethics in the humans who might create artificial agents).  We
begin to address this by presenting an embryonic formalization of a
key part of any virtue-ethics theory: namely, the learning of virtue
by a focus on exemplars of moral virtue.  Our work is based in part on
a computational formal logic previously used to formally model other
ethical theories and principles therein, and to implement these models
in artificial agents.
\end{abstract}

\section{Introduction}
\input{intro.tex}

\section{Why Virtuous Robots?}
\input{why.tex}

%%%%%%%%%%
\section{Surveying Virtue Ethics}

\input{related.tex}

\input{ve_overview_forinputing.tex}

%%%%%%%%%%

\section{Exemplarist Virtue Theory}

\input{evt.tex}

\section{The Goal}

\input{goal.tex}

\section{Building the Formalization}
\label{sect:calc}
\input{calculus.tex}

\subsection{Defining Admiration}
\label{sect:occ}
\input{occ.tex}

\label{sect:inf}
%% NAVEEN:  emotions.tex seems to be an empty file.  //S
%% SELMER: Removed

\subsection{Defining Traits }
\input{traits.tex}

\subsection{Defining Learning of Traits}
\input{learningTraits.tex}

\subsection{A Note on Learning Methods} 
\input{learning.tex}

\section{Defining Virtuous Person and Virtues}
\input{topdefs.tex}

\section{Implementation \& Simulation} 
\input{example.tex}

\section{Conclusion \& Future Work}

We have presented an initial formalization $\Vf$ of a virtue ethics
theory $\Vz$
%% NAVEEN: Shouldn't we refer to the specific label for this
%% particular virtue-ethics theory, in the overall family \VEE, to
%% connect to the notation used above?//S
%% SELMER: Done
in a calculus that has been used in automating other ethical
principles in deontological and consequentialist ethics.  Many
important questions have to be addressed in future research.  Among
them are questions about the nature and source of the utility
functions that are used in the definitions of emotions. Lacking in our
above model is an account of uncertainties and how they interact with
virtues. We plan to leverage an account of uncertainty for a fragment
of \DCEC\ presented in \cite{govindarajulu2017strength}. In future
work, we will compare learning traits with work on learning norms
\cite{sarathy2017learning}. The notion of learning we have presented
here is quite abstract. In order to handle more complex traits, more
sophisticated learning frameworks may have to be considered.  Finally,
we need to apply this model to more realistic examples and case
studies.
%% NAVEEN: I don't think the following sentence really adds anything.  //S
%
%% The lack of such formal examples and case studies is a bottleneck
%% here.
The way forward to the production of virtuous machines is thus
challenging, but we are confident that the foundation is now in place
for their eventual arrival.

%% SB***

\newpage \clearpage
\bibliographystyle{aaai}
\bibliography{main72,naveen}

\smallskip

\centerline{------}

\noindent 
\textbf{Acknowledgments}.  Support from ONR (to pursue morally
competent robots) and AFOSR (to pursue forms of logicist AI made
possible by high-expressivity calculi) has in part enabled the
research and engineering that underlies the present paper, and we are
most grateful.  We are grateful as well to three anonymous reviewers
for helpful comments on an eariler draft of the paper (but regret that
without additional space we couldn't address all their feedback).

\end{document}

%% file: commands.tex
\usepackage{times}
 \usepackage{soul}
\usepackage{paralist}

 \usepackage{amsfonts}
\usepackage{amsmath}
\usepackage{mathtools}    
\usepackage{relsize}
 \usepackage{proof}  
\usepackage{bussproofs}
\usepackage{xcolor}

\usepackage{pifont}% http://ctan.org/pkg/pifont
\usepackage{cleveref}

\newcommand{\IGNORE}[1]{}

\newcommand{\lsort}[1]{%
  \ensuremath{\mbox{\textsf{#1}}}}
\newcommand{\defsort}[2]{%
  \newcommand{#1}{\lsort{#2}}}
\defsort{\Action}{Action}
\defsort{\Time}{Time}
\defsort{\Self}{Self}

\defsort{\Agent}{Agent}
\defsort{\Entrant}{Entrant}
\defsort{\ActionType}{ActionType}
\defsort{\Moment}{Moment}
\defsort{\Boolean}{Formula}
\defsort{\PayOut}{PayOut}
\defsort{\Fluent}{Fluent}
\defsort{\Event}{Event}
\defsort{\Object}{Object}
\defsort{\RealTerm}{RealTerm}

\defsort{\Numeric}{Numeric}
\defsort{\Number}{Number}
\defsort{\Trolley}{Trolley}
\defsort{\Track}{Track}
\defsort{\Moveable}{Moveable}

\newcommand{\lsymbol}[1]{%
  \ensuremath{\mathit{#1}}}
\newcommand{\defsymbol}[2]{%
  \newcommand{#1}{\lsymbol{#2}}}
\defsymbol{\actionType}{action}
\defsymbol{\initially}{initially}
\defsymbol{\holds}{holds}
\defsymbol{\happens}{happens}
\defsymbol{\clipped}{clipped}
\defsymbol{\initiates}{initiates}
\defsymbol{\terminates}{terminates}
\defsymbol{\prior}{prior}
\defsymbol{\interval}{interval}
\defsymbol{\action}{action}

\defsymbol{\does}{does}
\defsymbol{\plans}{plans}
\defsymbol{\act}{act}
\defsymbol{\react}{react}
\defsymbol{\payTot}{pay_{tot}}
\defsymbol{\fight}{fight}
\defsymbol{\coop}{coop}
\defsymbol{\enter}{enter}
\defsymbol{\stayout}{stayout}
\defsymbol{\learns}{learns}
\defsymbol{\payoff}{payoff}
\defsymbol{\position}{position}
\defsymbol{\dead}{dead}
\defsymbol{\damaged}{damaged}

\defsymbol{\onrails}{onrails}
\defsymbol{\switch}{switch}
\defsymbol{\drop}{drop}

\newcommand{\lconstant}[1]{%
  \ensuremath{\mbox{\textsf{#1}}}}
\newcommand{\defconstant}[2]{%
  \newcommand{#1}{\lconstant{#2}}}
\defconstant{\Enter}{Enter}
\defconstant{\StayOut}{StayOut}
\defconstant{\Fight}{Fight}
\defconstant{\Acquiesce}{Acquiesce}
\defconstant{\cs}{cs }

\newcommand{\lmodality}[1]{%
  \ensuremath{\mathbf{#1}}}
\newcommand{\defmodality}[2]{%
  \newcommand{#1}{\lmodality{#2}}}
\defmodality{\common}{C}
\defmodality{\knows}{K}
\defmodality{\believes}{B}
\defmodality{\perceives}{P}
\defmodality{\mental}{M}

\defmodality{\desires}{D}
\defmodality{\intends}{I}
\defmodality{\says}{S}
\defmodality{\ought}{O}

\newcommand{\sep}{\ \lvert \ }

\usepackage{booktabs}
\usepackage{mdframed}
\usepackage{paralist}

\usepackage{color}

\newcommand{\DDE}{\ensuremath{\mathcal{{DDE}}}}
\newcommand{\DCEC}{\ensuremath{{\mathcal{{DCEC}}}}}

\newcommand{\type}[1]{\textsf{#1}}

\newcommand{\admires}{\ensuremath{\mathit{admires}}}
\newcommand{\exemplar}{\ensuremath{\mathit{Exemplar}}}
\newcommand{\trait}{\ensuremath{\mathbf{Trait}}}

%% file: intro.tex
What is virtue ethics?  One way of summarizing virtue ethics is to
contrast it with the two main families of ethical theories:
\textbf{deontological ethics} (\DE) and \textbf{consequentialism}
(\CE).  Ethical theories in the family \CE\ that are utilitarian in
nature hold that actions are morally evaluated based on their
\textbf{total utility} (or total \emph{dis}utility) to everyone
involved.  The best action is the action that has the highest total
utility.  In stark contrast, ethical theories in \DE\ emphasize
\textbf{inviolable principles}, and reasoning from those principles to
whether actions are obligatory, permissible, neutral,
etc.\footnote{Both the families \CE\ and \DE\ are crisply explained as
  being in conformity with what we say here in
  e.g.\ \cite{feldman_introductory_ethics}.}  In a departure from both
\DE\ and \CE, ethical theories in the virtue-ethics family \VE\ are
overall distinguished by the principle that the best action in a
situation, morally speaking, is the one that a \textbf{virtuous
  person} would perform \cite{intelligent_virtue}.  A virtuous person
is defined as a person who has learned and internalized a set of
habits or traits termed \textbf{virtuous}.  For a virtuous person,
virtuous acts become second-nature, and hence are performed in many
different situations, through time.

While there has been extensive formal and rigorous modeling done in
\DE\ and \CE, there has been little such work devoted to formalizing
and mechanizing \VE.  Note that unlike \DE\ and \CE, it is not
entirely straightforward how one could translate the concepts and
principles in \VE\ into a form that is precise enough to be realized
in machines.  Proponents of \VE\ might claim that it is not feasible
to do so given \VE's emphasis on persons and traits, rather than
individual actions or consequences.  From the perspective of machine
ethics, this is not satisfactory.  If \VE\ is to be on equal footing
with \DE\ and \CE\ for the purpose of building morally competent
machines, AI researchers need to start formalizing parts of virtue
ethics, and to then implement such formalization in computation.

% Commenting this for now.
% (After all, machines don't yet understand that which is
% informal; witness e.g.\ SIRI.)

We present one such formalization herein; one that uses learning and
is based on a virtue-ethics theory presented by Zagzebski
\cite{zagzebski2010exemplarist}.  The formalization is presented
courtesy of an expressive computational logic that has been used to
model principles in both \CE\ and \DE\ [e.g.\
\cite{nsg_sb_dde_2017,dde_self_sacrifice_2017}].\footnote{See
  \cite{sb_etal_ieee_robots} for an introduction to the logicist
  methodology for building ethical machines. }  The formalization
answers, abstractly, the following two questions:
\begin{footnotesize}
\begin{mdframed}[linecolor=white, frametitle=Questions,
  frametitlebackgroundcolor=gray!10, backgroundcolor=gray!05,
  roundcorner=8pt]
  \begin{enumerate}
  \item[($\mathbf{Q_1}$)] When can we say an agent is virtuous?
  \item[($\mathbf{Q_2}$)] What is a virtue?
%%  \item[$\mathbf{Q_3}$:] How can an agent become virtuous?
  \end{enumerate}
\end{mdframed}
\end{footnotesize}
The plan for the paper is as follows.  First, we briefly look at why
virtuous machines might be useful, and then we briefly cover related
work that can be considered as formalizations of virtue ethics.  Next,
we present an overview of virtue ethics itself, and specifically show
that an emphasis on moral exemplars makes good sense for any attempt
to engineer a virtuous machine.  We next present one version
of virtue ethics, $\VE_z$ (Zagzebski's version of virtue ethics), that
we seek to formalize fully.  Then, our calculus and the formalization
itself (\Vf) are presented.  We conclude by discussing future work and
remaining challenges.

% Given any formalization of any ethical theory, we can then ask two
% validation questions: ($\mathbf{V_1}$): Does the formalization model
% the ethical theory?, and ($\mathbf{V_2}$): Is the formalization
% useful?  We also present initial work addressing these two validation
% questions.

%%% Local Variables:
%%% mode: latex
%%% TeX-master: "main"
%%% End:

%% file: why.tex
Note that we do not advocate that machine ethicists pursue virtue
ethics over other familiies of ethical theories.  Our goal in the
present paper is merely to formalize one version of virtue ethics
within the family \VE.  That said, why might virtue ethics be
considered over consequentialism or deontological ethics for building
morally competent machines?  To partially answer this question, we
take a short digression into a a series of conditions laid out by
\citeauthor{alfano2013identifying}, and characterized as identifying
the core of virtue ethics:

\begin{small}
  \begin{mdframed}[linecolor=white, frametitle= Hard Core of
  Virtue Ethics (quoting 
  \cite{alfano2013identifying}), frametitlebackgroundcolor=gray!08,
  backgroundcolor=gray!05, roundcorner=8pt]

\begin{enumerate}[(1)]
\item \textbf{acquirability} It is possible for a non-virtuous person
  to acquire some of the virtues.
\item \textbf{stability} If someone possesses a virtue at time $t_1$,
  then \textit{ceteris paribus} she will possess that virtue at a
  later time $t_2$.
\item \textbf{consistency} If someone possesses a virtue sensitive to
  reason $r$, then \emph{ceteris paribus} she will respond to $r$ in most
  contexts.
\item \textbf{access} It is possible to determine what the virtues
  are.
\item \textbf{normativity} \emph{Ceteris paribus}, it is better to
  possess a virtue than not, and better to possess more virtues than
  fewer.
\item \textbf{real saints}  There is a non-negligible cohort of saints
  in the human population.
\item \textbf{explanatory power} If someone possesses a virtue, then
  reference to that virtue will sometimes help to explain her
  behavior.
\item \textbf{predictive power} If someone possesses a high-fidelity
  virtue, then reference to that virtue will enable nearly certain
  predictions of her behavior; if someone possesses a low-fidelity
  virtue, then reference to that virtue will enable weak predictions
  of her behavior.
\item \textbf{egalitarianism} Almost anyone can reliably act in
  accordance with virtue.
\end{enumerate}
\end{mdframed}
\end{small}

Particularly, we feel that if the conditions of \textbf{stability},
\textbf{consistency}, \textbf{explanatory power}, and
\textbf{predictive power} hold, then virtuous agents or robots might
be easier for humans to understand and interact with (compared to
consequentialist or deontological agents or robots).  This is but our
initial motivation; we now present an overview of virtue ethics, in
order to show that our focus specifically on learning of virtuous
behavior from moral exemplars is advisable.

%%% Local Variables:
%%% mode: latex
%%% TeX-master: "main"
%%% End:

%% file: related.tex
\begin{figure}[h!]
 \centering
 {
  \includegraphics[width=0.44\linewidth]{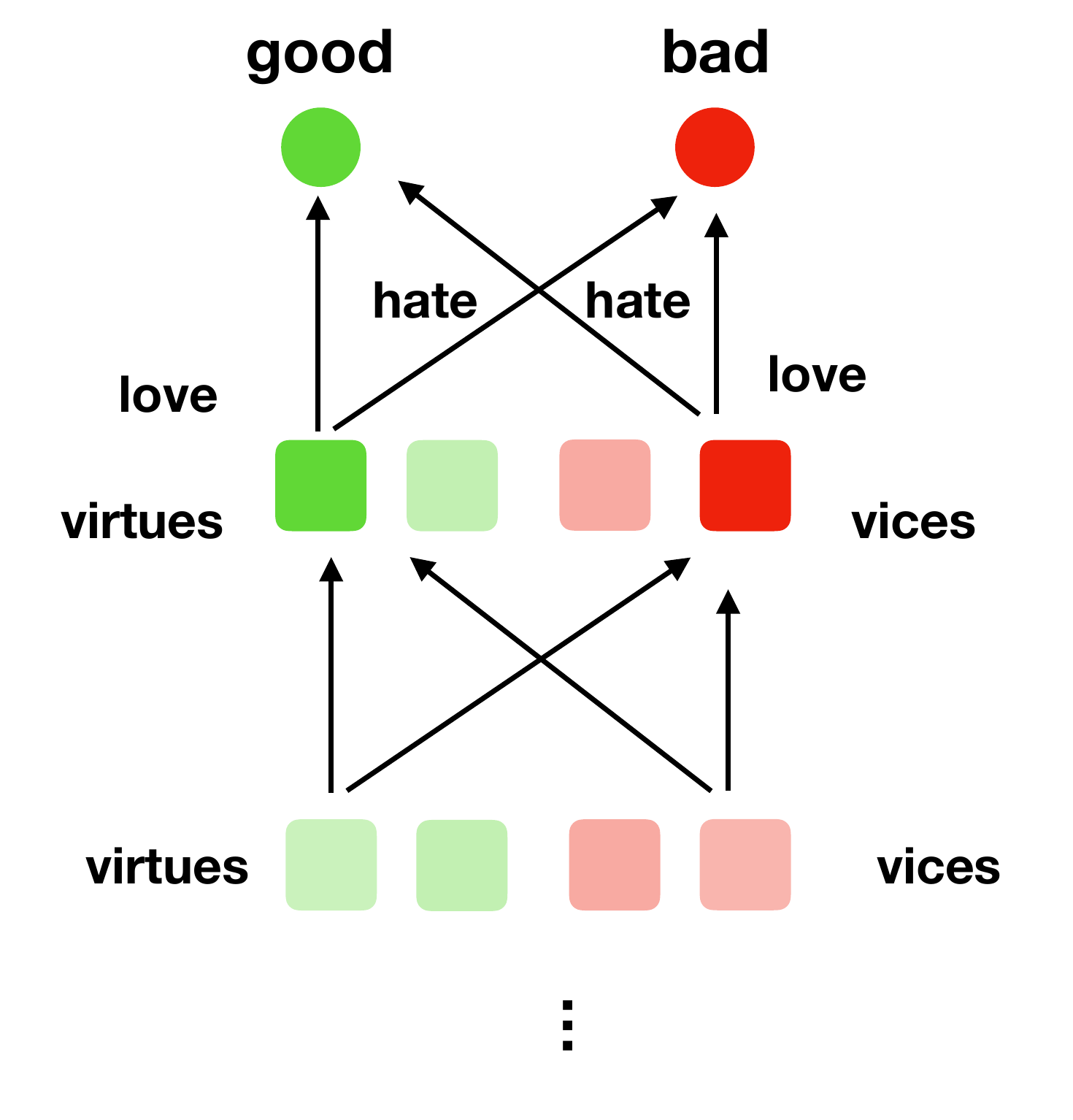}}
\caption{\textbf{Hurka's Account} Virtues (vices) are defined
  recursively as love of good (bad) states-of-affairs or hate (love)
  of bad states of affairs.}
 \label{fig:ccOverview}
\end{figure}

See \cite{scheutzmallemoral} for a general introduction to the field
of moral robots. We begin our survey by reporting that Hurka
\cite{hurka2000virtue} presents an ingenious formal account involving
a recursive notion of goodness and badness.  The account starts with a
given set of primitive good and bad states-of-affairs.  Virtues are
then defined as love of good states-of-affairs or hatred of bad
states-of-affairs.  Vice is defined as love of bad states-of-affairs
or hatred of good states-of-affairs.  Virtues and vices are then
themselves taken to be good and bad states-of-affairs, resulting in a
recursive definition (see Figure~\ref{fig:ccOverview}) that is
attractive to AI researchers and computer scientists.  But despite
this, and despite our sense that the main problems with Hurka's
account are rectifiable \cite{hiller2011unusual}, we feel that Hurka's
definition might not capture central aspects of virtue
\cite{miles2013against}.  More problematic is that it must be shown
that Hurka's account is different from rigorous and formal accounts of
\CE, which after all are themselves invariably based upon good and bad
states-of-affairs.  Moreover, it is not clear to us how Hurka's
account is amenable to automation.  Therefore, we now proceed to step
back and survey the overarching family \VE\ of virtue ethics, to
specifically pave a more promising AI road: a focus on moral
exemplars.
 
%%++++++++++++++++++++++++++++++++++++++++++++++++++++++++++++++++++++
%%++++++++++++++++++++++++++++++++++++++++++++++++++++++++++++++++++++

%%% Local Variables:
%%% mode: latex
%%% TeX-master: "main"
%%% End:

%% file: ve_overview_forinputing.tex
\subsection{Virtue Ethics:  Overview to Exemplarism}
\label{subsect:ve_over_exemplarism}

%% \subsection{\VE\ is Somewhat Unfamiliar to Folks}
%
%% Explain that VE is a bit unfamiliar, in context
  The core concepts of consequentialist ethical theories (i.e.\ of
  members of \CE), at least certainly in the particular such theory
  known as \textit{utilitarianism}, are doubtless at minimum
  relatively familiar to most of our readers.  For instance, most in
  our audience will know that utilitarianism's core tenet is that
  actions are obligatory just in case they have the consequence of
  maximizing happiness, and are forbidden exactly when they fail to so
  maximize.  A parallel state-of-affairs holds for at least basic
  knowledge of deontological ethical theories (= family \DE): most
  readers have for instance some familiarity with Kant's moral system
  in family \DE, and specifically with his famous ``categorical
  imperative,'' which, paraphrasing, says that, unconditionally, one
  must always act in such a way that this behavior could be
  universalized.\footnote{This imperative is first set out in --- as
    it's known in abbreviation --- \textit{Groundwork}; see
    \cite{practical_philosophy_with_groundwork}.  It's generally
    thought by ethicists, and this may be convenient for machine/AI
    ethics, that Kant had in mind essentially a decision procedure to
    follow in the attempt to behave in an ethically correct manner.
    For a lucid and laconic overview of this point, see
    \cite{sep_kant_moral_philosophy}; and
    cf.\ \cite{powers_kantian_machine}.} In addition, generally people
  are familiar with the core tenet of divine-command ethical theories
  (i.e.\ of members of \DCE), which is (approximately) that actions
  are obligatory for humans if and only if God commands that these
  actions be performed (a particular member of \DE is specified in
  \cite{quinn_divine_command}).  However, in our experience the
  epistemic situation is radically different when it comes to the
  family of ethical theories \textit{virtue ethics} (= \VE).  For
  while it's true that generally educated people can be assumed to be
  acquainted with the concept of virtue, and with many things long
  deemed to be virtues (e.g.\ bravery), an understanding of virtue
  ethics at the level of ethical \emph{theory} cannot be assumed.  We
  therefore now provide a rapid (and admittedly cursory) synopsis of
  \VE, by drawing from \cite{vallor_ve_book}, and to some degree from
  \cite{intelligent_virtue}.  It will be seen that \VE\ makes central
  use of exemplars, and of learning and development that revolves
  around them.  Hence we shall arrive at a convenient entry point for
  our AI work devoted to trying to design and build a virtuous
  machine.
%% NAVEEN: One could go through the paper and opt to use only
%% `machine' or only `robot.'  I'll leave it to you to decide, but at
%% this point I changed the title to refer to `Machine.'  See my
%% comment near \title.  Thx.  //S

%% \subsection{Vallor's Affirmation of Exemplars and Learning}
%
%% Extraction from Vallor's book:
  Obviously we cannot in the span of the space we have at hand do full
  justice to the book-length treatment of \VE\ that is
  \cite{vallor_ve_book}.  But we can quickly establish that our
  technical work, in its focus on the cultivation of virtue for a
  machine via learning from exemplars, is not merely based on a
  single, idiosyncratic member of \VE, and on one peripheral aspect of
  this member.  On the contrary, study of the work of Vallor and other
  scholars concerned with a characterization of the family
  \VE\ confirms that our exploitation specifically of Zagzebski's
  \cite{zagzebski2010exemplarist} focus, from the standpoint of the
  field of ethics itself, is a worthy point of entry for AI
  researchers.

  To begin, Vallor, drawing on and slightly adapting Van Norden's
  \cite{van_norden_ve_consequentialism} sets out a quartet of
  commonalities that at least seem to be true of all members of \VE,
  and the second one is: ``A conception of moral virtues as cultivated
  states of character, manifested by those exemplary persons who have
  come closest to achieving the highest human good'' (\P 5, \S
  2.2).\footnote{In her book, Vallor gives her own more detailed and
    technologically relevant list of seven core elements that can be
    viewed as common to all members of \VE\ (or two what she refers to
    as ``traditions'' within virtue ethics).  We do not have the space
    to discuss this list, and show that it fits nicely with our
    technical work's emphasis on exemplars and learning therefrom.}
  But given our specific efforts toward engineering a virtuous
  machine, it is important to note that Vallor specifically informs us
  about the key concepts of exemplars in the particular members of the
  \VE family; to pick just one of many available places, she writes:

    \begin{small}
    \begin{quote}
      Buddhism's resonances with other classical virtue traditions do
      not end here.  As with the central role granted by Confucian and
      Aristotelian ethics to `exemplary persons' (the \textit{junzi}
      and \textit{phronimoi} respectively), \textit{bodhisattvas}
      (persons actively seeking enlightenment) generally receive
      direction to or assistance on the path of self-cultivation from
      the community of exemplary persons to which they have access.
      In Buddhism this is the monastic community and lay members of
      the \textit{Sangha} $\ldots$ [\P 5, \S 2.1.3,
        \cite{vallor_ve_book}]
    \end{quote}
    \end{small}

%% \subsection{Annas's Book Fits Too}
%
%% What about Annas's book?  Here we go:
%
  We said above that we would also draw, albeit briefly, from a second
  treatment of \VE, viz.\ \cite{intelligent_virtue}, in order to pave
  the way into our AI-specific, exemplar-based technical work.  About
  this second treatment we report only that it is one based squarely
  on a ``range of development'' (\P 3, \S Right Action in Ch.\ 3),
  where the agent (a human in her case) gradually develops into a
  truly virtuous person, beginning with unreflective adoption of
  direct instruction, through a final phase in which ``actions are
  based on understanding gained through experience and reflection''
  (ibid.).  Moreover, Annas explicitly welcomes the analogy between an
  agent's becoming virtuous, and an agent's becoming, say, an
  excellent tennis-player or pianist.  The idea behind the similarity
  is that ``two things are united: the \emph{need to learn} and the
  \emph{drive to aspire} (emphasis hers; \P 4 Ch.\ 3).  In addition,
  following Aristotle on \VE\ (e.g.\ see
  \cite{aristotle_nicomachean_ethics} 1103), no one can become a
  master tennis-player or pianist without, specifically, playing
  tennis/the piano with an eye to the mastery of great exemplars in
  these two domains.

%% \subsection{Transition \P}
%
%% Transition paragraph:
  In order to now turn to specific AI work devoted to engineering a
  virtuous machine, we move from completed consideration of the general
  foundation of \VE, and its now-confirmed essential use of moral
  exemplars, to a specific use of such exemplars that appears ripe for
  mechanization.

%% file: evt.tex
\textbf{Exemplarist virtue theory} (\Vz) builds on the \textbf{direct
  reference theory} (DRT) of semantics.  Briefly, in DRT, given a word
or term $w$, its meaning $\mu(w)$ is determined by what the word
points out, say $p$, and not by some definition $d$.  For example, for
a person to use the word \emph{``water,''} in a correct manner, that
person neither needs to possess a definition of water nor needs to
understand all the physical properties of water.  The person simply
needs to know which entity the word \emph{``water''} picks out in
common usage.

In \Vz, persons understand moral terms, such as \emph{``honesty,''},
in a similar manner.  That is, moral terms are understood by persons
through direct references instantiated in \textbf{exemplars}.  Persons
identify moral examplars through the emotion of \textbf{admiration}.
The emotions of admiration and contempt play a foundational role in
this theory. \Vz\ posits a process very similar to scientific or
empirical investigation.  Exemplars are first identified and their
traits are studied; then they are continously further studied to
better understand their traits, qualities, etc.  The status of an
individual as an exemplar can change over time.  Below is an informal
version that we seek to formalize:

\begin{footnotesize}

\begin{mdframed}[frametitle={ Informal Version \Vz},linewidth=0,backgroundcolor=gray!05,frametitlebackgroundcolor=gray!10]

\begin{enumerate}
\item[$\mathbf{I}_1$] Agent or person $a$ perceives a person $b$
  perform an action $\alpha$.  This observation causes the emotion of
  admiration in $a$.

\item[$\mathbf{I}_2$] $a$ then studies $b$ and seeks to
  learn what traits (habits/dispositions) $b$ has.
\end{enumerate}

\end{mdframed}
\end{footnotesize}

%%% Local Variables:
%%% mode: latex
%%% TeX-master: "main"
%%% End:

%% file: goal.tex
From the above presentation of \Vz, we can glean the following
distilled requirements that should be present in any formalization.

\begin{small}

  \begin{mdframed}[frametitle={$\Vf$ Formalization
      Components},linewidth=0,backgroundcolor=gray!05,frametitlebackgroundcolor=gray!10]

\begin{enumerate}
\item[($\mathbf{R}_1$)] A formalization of emotions, particularly admiration.
\item[($\mathbf{R}_2$)] A representation of traits.
%% NAVEEN: I find `notion' here indeterminate.  Would it be
%% better/more informative to say something like: A representation of
%% traits. ?  //S
\item[($\mathbf{R}_3$)] A process of learning traits (and not just
  simple individual actions) from a small number of observations.
\end{enumerate}

\end{mdframed}
\end{small}

%%% Local Variables:
%%% mode: latex
%%% TeX-master: "main"
%%% End:

%% file: calculus.tex
For fleshing out the above requirements and formalizing \Vz, we will
use the \textbf{deontic cognitive event calculus} (\DCEC), a
computational formal logic.  This logic was used previously in
\citelow{nsg_sb_dde_2017,dde_self_sacrifice_2017} to automate versions
of the Doctrine of Double Effect (\DDE), an ethical principle with
deontological and consequentialist components.  \DCEC\ has also been
used to formalize \textit{akrasia} (the process of succumbing to
temptation to violate moral principles) \cite{akratic_robots_ieee_n}.
Fragments of \DCEC\ have been used to model highly intensional
reasoning processes, such as the false-belief task
\citelow{ArkoudasAndBringsjord2008Pricai}.\footnote{\DCEC\ is both
  \emph{intensional} and \emph{intentional}. There is a difference
  between {intensional} and intentional systems.  Broadly speaking,
  extensional systems are formal systems in which the references and
  meanings of terms are independent of any context.  Intensional
  systems are formal systems in which meanings of terms are dependent
  on context, such as the cognitive states of agents, time, etc.
  Modal logics used for modeling beliefs, desires, and intentions are
  considered intensional systems.  Please see the appendix in
  \citelow{nsg_sb_dde_2017} for a more detailed discussion.}

\subsection{Overview of \DCEC}

\DCEC\ is a quantified multi-operator\footnote{The full catalogue of
  available operators exceeds those for belief, desire, and intention,
  and \textit{a fortiori} exceeds the available operators in any
  standard modal logic designed to formalize e.g.\ only either
  alethic, epistemic, or deontic phenomena.} modal logic (also known
as sorted first-order multi-operator modal logic) that includes the
event calculus, a first-order calculus used for commonsense reasoning
over time and change \cite{mueller_commonsense_reasoning_2nd_ed}.
This calculus has a well-defined syntax and proof calculus; see
Appendix A of \citelow{nsg_sb_dde_2017}.  The proof calculus is based
on natural deduction
\citelow{gentzen_investigations_into_logical_deduction}, and includes
all the introduction and elimination rules for first-order logic, as
well as inference schemata for the modal operators and related
structures.  As a sorted calculus, \DCEC\ can be regarded analogous to
a typed programming language.  We show below some of the important
sorts used in \DCEC.  Among these, the \type{Agent}, \type{Action},
and \type{ActionType} sorts are not native to the event calculus.\\

% \begin{footnotesize}
% \begin{mdframed}[frametitle= {\fontencoding{U}\fontfamily{futs}\selectfont\char 66\relax} Modeling Knowlege in First-order
%   Logic , frametitlebackgroundcolor=red!25,linecolor=white,backgroundcolor=orange!05]
% \begin{prooftree}
% \AxiomC{$\mathbf{K}\left(a,\  \mathsf{Murderer}(\mathit{owner}(knife))\right)$}
% \AxiomC{$ \mathit{adam} =  \mathsf{Murderer}(\mathit{owner}(knife))$}
% \BinaryInfC{$ \mathbf{K}\left(a,\ \mathit{x}= \mathit{y}\right)$}
% \AxiomC{$\lnot \mathit{Knows}\left(a,\ \mathit{x} = \mathit{y}\right)$}
% \BinaryInfC{$\phi \land \lnot \phi$}
% \end{prooftree}
% \end{mdframed}
% \end{footnotesize}
% \vspace{0.1in}

% A classic introduction to intensional logic is Zalta's
%   \shortcite{zalta1988intensional}, which goes so far as to allow that
%   agents can have beliefs about physically and mathematically
%   impossible objects. \IGNORE{ In some of our cognitive calculi, up to 33
%   separate emotional states are represented using our array of
%   intensional operators.}

\begin{scriptsize}
\begin{tabular}{lp{5.8cm}}  
\toprule
Sort    & Description \\
\midrule
\type{Agent} & Human and non-human actors.  \\

\type{Time} & The \type{Time} type stands for time in the domain;
e.g.\ simple, such as $t_i$, or complex, such as
$birthday(son(jack))$. \\

 \type{Event} & Used for events in the domain. \\
 \type{ActionType} & Action types are abstract actions.  They are
  instantiated at particular times by actors.  Example: eating.\\
 \type{Action} & A subtype of \type{Event} for events that occur
  as actions by agents. \\
 \type{Fluent} & Used for representing states of the world in the
  event calculus. \\
\bottomrule
\end{tabular}
\end{scriptsize} \\

Note: actions are events that are carried out by an agent.  For any
action type $\alpha$ and agent $a$, the event corresponding to $a$
carrying out $\alpha$ is given by $\action(a, \alpha)$.  For instance,
if $\alpha$ is \textit{``running''} and $a$ is \textit{``Jack'' },
$\action(a, \alpha)$ denotes \textit{``Jack is running''}.

\subsubsection{Syntax}
\label{subsect:syntax}

The syntax has two components: a first-order core and a modal system
that builds upon this core.  The figures below show the formal
language and inference schemata of \DCEC.
%% NAVEEN: I don't think we need the following sentence.  It's
%% redundant.  Also, I've referred here (just above) to the formal
%% language and/vs the inference schemata.
%
%% The syntax is quantified modal logic.
% The first-order core of \DCEC\ is the \emph{event calculus}
% \cite{mueller_commonsense_reasoning}. 
Commonly used function and
relation symbols of the event calculus are included.  Any formally
defined calculi (e.g.\ the venerable \emph{situation calculus}) for
modeling commonsense and physical reasoning can be easily switched out
in-place of the event calculus.

% Since the semantics of \DCEC\ is proof-theoretic, as long as a new
% construct has appropriate inference schemata, the extension is
% sanctioned.

The modal operators present in the calculus include the standard
operators for knowledge $\knows$, belief $\believes$, desire
$\desires$, intention $\intends$, obligation $\ought$ etc.  For
example, consider $\believes\left(a, t, \phi\right)$, which says that
agent $a$ believes at time $t$ the proposition $\phi$.  Here $\phi$
can in turn be any arbitrary formula.

 \begin{scriptsize}
\begin{mdframed}[linecolor=white, frametitle=Syntax (fragment),
  frametitlebackgroundcolor=gray!10, backgroundcolor=gray!05,
  roundcorner=8pt]
 \begin{equation*}
 \begin{aligned}
    \mathit{S} &::= 
    \begin{aligned}
      & \Agent \sep \ActionType \sep \Action \sqsubseteq
      \Event \sep \Moment  \sep \Fluent \\
    \end{aligned} 
    \\ 
    \mathit{f} &::= \left\{
    \begin{aligned}
      & \action: \Agent \times \ActionType \rightarrow \Action \\
      &  \holds: \Fluent \times \Moment \rightarrow \Boolean \\
      & \happens: \Event \times \Moment \rightarrow \Boolean \\
      & \prior: \Moment \times \Moment \rightarrow \Boolean\\
    \end{aligned}\right.\\
        \mathit{t} &::=
    \begin{aligned}
      \mathit{x : S} \sep \mathit{c : S} \sep f(t_1,\ldots,t_n)
    \end{aligned}
    \\ 
    \mathit{\phi}&::= \left\{ 
    \begin{aligned}
     & q:\Boolean \sep  \neg \phi \sep \phi \land \psi \sep \phi \lor
     \psi \sep \forall x: \phi(x) \sep \\\
 &\perceives (a,t,\phi)  \sep \knows(a,t,\phi) \sep     \\ 
& \common(t,\phi) \sep
 \says(a,b,t,\phi) 
     \sep \says(a,t,\phi) \sep  \believes(a,t,\phi) \\
 & \ought(a,t,\phi,(\lnot)\happens(action(a^\ast,\alpha),t'))
      \end{aligned}\right.
  \end{aligned}
\end{equation*}
\end{mdframed}
\end{scriptsize}

\subsubsection{Inference Schemata}

The figure below shows a fragment of inference schemata for \DCEC.
$I_\mathbf{B}$ is an inference schema that let us model idealized
agents that have their knowledge and belief closed under the \DCEC\
proof theory.  While normal humans are not deductively closed, this
lets us model more closely how deliberative agents such as organizations
and more strategic actors reason.  (Some dialects of cognitive calculi
restrict the number of iterations on intensional operators.)  $I_{12}$
states that if an agent $s$ communicates a proposition $\phi$ to $h$,
then $h$ believes that $s$ believes $\phi$.  $I_{14}$ dictates how
obligations propagate to intentions.

\begin{scriptsize}
\begin{mdframed}[linecolor=white, frametitle=Inference Schemata (fragment), frametitlebackgroundcolor=gray!10, backgroundcolor=gray!05, roundcorner=8pt]
\begin{equation*}
\begin{aligned}
&\infer[{[I_{\believes}]}]{\believes(a,t_2,\phi)}{\believes(a,t_1,\Gamma), \ 
    \ \Gamma\vdash\phi, \ \ t_1 < t_2}  \hspace{12pt}\infer[{[I_{12}]}]{\believes(h,t,\believes(s,t,\phi))}{\says(s,h,t,\phi)}
\\
&\infer[{[I_{14}]}]{\knows(a,t,\intends(a,t,\chi))}{\begin{aligned}\ \ \ \ \believes(a,t,\phi)
 & \ \ \
 \believes(a,t,\ought(a,t,\phi, \chi)) \ \ \ \ought(a,t,\phi,
 \chi)\end{aligned}}
\end{aligned}
\end{equation*}
\end{mdframed}
\end{scriptsize}

We also define the following inference-schemata-based relationships
between expressions in our calculus.

\noindent \textbf{Generalization of Formulae}. The generalization of a
set of formulae $\Psi$, is a set of formulae $\Phi$ from which any
element of $\Psi$ can be inferred: $\Phi\vdash \bigwedge \Psi$. This
is denoted by $g(\Psi) = \Phi$.

\noindent \textbf{Generalization of Terms}: A term $x$ is a
generalization of a term $y$ if given any first-order predicate $P$,
from $P(x)$ we can derive $P(y)$: $\{P(x)\} \vdash P(y)$. This
is denoted by $g(y) = x$.

\subsection{Semantics}
\DCEC\ uses \textit{proof-theoretic} semantics
\cite{gentzen_investigations_into_logical_deduction,proof-theoretic_semantics_for_nat_lang},
an approach commonly associated with natural deduction inference
systems.  Briefly, in this approach, meanings of modal operators are
defined via functions over proofs. Specifying semantics then reduces
to specifying inference schemata.

\subsection{Events, Fluents, and Utilities}
\input{utility.tex}

With the calculus given above, we now move on to specifying parts of
the formalization $\Vf$, that is, $\mathbf{R}_1$, $\mathbf{R}_2$, and
$\mathbf{R}_3$.

 % Extant first-order
% modal logic theorem provers that can work with arbitrary inference
% schemata are built upon first-order theorem provers.  They achieve the
% reduction to first-order logic via two methods.  In the first method,
% modal operators are simply represented by first-order predicates. This
% approach is the fastest but can quickly lead to well-known
% inconsistencies as demonstrated in
% \citelow{selmer_naveen_metaphil_web_intelligence}. In the second method,
% the entire proof theory is implemented intricately in first-order
% logic, and the reasoning is carried out within first-order logic.
% Here, the first-order theorem prover simply functions as a declarative
% programming system.  This approach, while accurate, can be
% excruciatingly slow.  We use a different approach, in which we
% alternate between calling a first-order theorem prover and applying
% modal inference schemata.  When we call the first-order prover, all
% modal atoms are converted into propositional atoms (i.e., shadowing),
% to prevent substitution into modal contexts.  This approach achieves
% speed without sacrificing consistency.  The prover also lets us add
% arbitrary inference schemata to the calculus by using a
% special-purpose language.  
 
%
%% \color{blue}
%
%%% Local Variables:
%%% mode: latex
%%% TeX-master: "main"
%%% End:

%
%% \color{blue}
%
%%% Local Variables:
%%% mode: latex
%%% TeX-master: "main"
%%% End:

%% file: utility.tex
In the event calculus, fluents represent states of the world.  Our
formalization of admiration requires a notion of utility for states of
the world.  Therefore, we assign utilities to fluents through a
utility function: $\mu: \Fluent \times \Time \rightarrow \mathbb{R}
$. An event can initiate one or more fluents. Therefore, events can
also have a utility associated with them.  For an event $e$ at time
$t$, let $e_I^{t}$ be the set of fluents initiated by the event, and
let $e_T^{t}$ be the set of fluents terminated by the event.  If we
are looking up till horizon $H$, then ${\nu}(e, t)$, the total utility
of event $e$ at time $t$, is:
\begin{footnotesize}
\begin{equation*}
\begin{aligned}
{\nu}(e, t) =
\mathlarger{\mathlarger{\sum}}_{y=t+1}^{H}\Bigg(\sum_{f\in {e_I^t}}
\mu(f,y) - \sum_{f\in {e_T^t}} \mu(f,y)\Bigg)
\end{aligned}
\end{equation*}
 \end{footnotesize}
With the calculus given above, we now move on to specifying parts of
the formalization $\Vf$, that is, $\mathbf{R}_1$, $\mathbf{R}_2$, and
$\mathbf{R}_3$.

%%% Local Variables:
%%% mode: latex
%%% TeX-master: "main"
%%% End:

%% file: occ.tex
We start with $\mathbf{R}_1$ and formalize admiration in \DCEC.  To
acheive this, we build upon the \textbf{OCC model}.  There are many
models of emotion from psychology and cognitive science.  Among these,
the OCC model \cite{occ_main} has found wide adoption among computer
scientists.  Note that the model presented by \cite{occ_main} is
informal in nature and one formalization of the model has been
presented in \cite{adam2009logical}.  This formalization is based on
propositional modal logic, and while comprehensive and elaborate, is
not expressive enough for our modelling, which requires at the least
quantification over objects.

In OCC, emotions are short-lived entities that arise in response to
\emph{events}.
%% NAVEEN: I don't understand the previous sentence.  //S
%% SELMER: Fixed.
Different emotions arise based
on: \begin{inparaenum}[(i)] \item whether the \emph{consequences} to
  events are positive (desirable) or negative (undesirable); \item
  whether the event has occured; and \item whether the event has
  consequences for the agent or for another agent. \end{inparaenum}
OCC assumes an undefined primitive notion of an agent being
\emph{pleased} or \emph{displeased} in response to an event.  We
represent this notion by a predicate $\Theta$ in our formalization.
In OCC, admiration is defined as ``\emph{(approving of) someone else’s
  praiseworthy action}.''  We translate this definition into \DCEC\ as
follows.  An agent $a$ is said to admire another agent $b$'s action
$\alpha$, if agent $a$ believes the action is a good action.  An
action $\action(b, \alpha)$ is a considered a good action if $ {\nu}(
\actionType(b, \alpha), t) > 0$.  In OCC, agents can admire only other
agents and not themselves.  This is captured by the inequality
$a\lnot=b$

\begin{footnotesize}
\begin{mdframed}[linecolor=white,nobreak, frametitle=($\mathbf{R}_1$) Admiration in \DCEC,
  backgroundcolor=gray!05, frametitlebackgroundcolor=gray!10,
  roundcorner=8pt]
\vspace{-0.13in}
\begin{equation*}
\begin{aligned}
  &\ \  \ \ \ \  \ \  \ \  \ \ \ \  \ \  \ \ \ \  \  \ \ \
  \holds(\admires(a,b, \alpha), t)\\
 & \ \  \ \ \ \  \ \  \ \  \ \ \ \  \ \  \ \ \ \  \  \ \ \  \ \ \ \ \
 \ \ \ \ \  \ \  \leftrightarrow\\
 & \color{gray!50}\left(\color{black}\begin{aligned} &  \hspace{0.35\linewidth}
    \mathlarger{\mathlarger\Theta(a,t')}   \hspace{0.05\linewidth}\mathlarger{\mathlarger{\mathlarger\land}} \\ \ &\believes\left(a, t,
       \left[\begin{aligned} & (a \not= b) \land
          (t' < t) \\ & \land \happens(\action(b, \alpha), t') \land\\ & {\nu}(
           \actionType(b, \alpha), t) > 0\end{aligned}\right]\right) \end{aligned}\color{gray!50}\right)\color{black}
\end{aligned}
\end{equation*}

\end{mdframed}
 
\end{footnotesize}

% \begin{center}
% \begin{footnotesize}
% \begin{tabular}{llcl}  
% \toprule
% \textbf{Emotion Type} & \textbf{Response} &  \textbf{Agent}  & \textbf{Consequences}  \\
% \midrule
% \ Joy & Pleased & Self & Desirable \\
%  \Distress & Displeased & Self & Undesirable \\
% \midrule
%  \HappyFor & Pleased & Other & Desirable \\
% \Gloating  & Pleased & Other & Undesirable \\
%  \PityFor & Displeased & Other & Undesirable \\
%  \Resentment & Displeased & Other & Desirable \\
% \bottomrule
% \end{tabular}
% \end{footnotesize}
% \end{center}

% As can be in the above table, among other things, our
% formalization needs to distinguish between the self and other agents
% in a non-trivial manner.

% \begin{description}
% \item[Joy] Pleased in response to an event with positive cosequences has occured
% \item[Distress] Pleased in response to an with negative cosequences has occured
% \item[Happy-For] Pleased in response to an event with positive cosequences for another agent has occured
% \item[Pity-For] Displeased in response to an event with negative cosequences for another agent has
%   occured
% \item[Gloating] Pleased in response to an event with negative cosequences for another agent has occured
% \end{description}

%%% Local Variables:
%%% mode: latex
%%% TeX-master: "main"
%%% End:

%% file: traits.tex
To satisfy $\mathbf{R}_2$, we need to define traits.  We define a \emph{situation}
$\sigma(t)$ as simply a collection of formulae that describe what
fluents hold at a time $t$, along with other event-calculus
constraints and descriptions (sometimes we use $\sigma(t)$ to represent
the conjunction of all the formulae in $\sigma(t)$.)
% An action type $\alpha$ is said to be an \emph{alternative} in a
% situation $\sigma(t)$ for an agent $a$ at time $t$ if the following
% holds:

% $$\sigma(t)+
% \happens\big(\action(\alpha, a), t\big) \not\vdash \bot$$

% Let $\Pi_{\sigma(t)}$ denote the set of all the alternatives in a
% situation $\sigma(t)$.

\begin{footnotesize}

  \begin{mdframed}[linecolor=white, frametitle= ($\mathbf{R}_2$)
    Trait, frametitlebackgroundcolor=gray!10, backgroundcolor=gray!05,
    roundcorner=8pt] An agent $a$ has a situation $\sigma$ and action
    type $\alpha$ as an $m$-\emph{trait}
    $\langle \sigma, \alpha \rangle$ if there are at least $m$
    situations $\{ \sigma_1 , \sigma_2 , \ldots, \sigma_m
    \}$ % there are
  % $n = \sum^m_{i=1} \Pi_{\sigma(i)}$ available alternatives in all the
  % situations but in which \emph{instantiations} of $\alpha$ are
  in which \emph{instantiations} of $\alpha$ are performed, and
  $\sigma$ is the generalization of the situations. 

\end{mdframed}
\end{footnotesize}

A trait $\langle \sigma, \alpha \rangle$ can be represented as single
formula:
$$\tau \equiv  \sigma \land happens(\action(\alpha, a), t)$$
We introduce a new modal operator $\trait$ that can then be applied to
the collection of formulae $\tau$ denoting a trait. $\trait(\tau, a)$
says that agent $a$ has trait $\tau$. The following inference schema
then applies to $\trait$:

\begin{footnotesize}
\begin{mdframed}[linecolor=white, frametitle= ($\mathbf{R}_2$)  Inference Schema for $\trait$,
  frametitlebackgroundcolor=gray!10, backgroundcolor=gray!05,
  roundcorner=8pt]
  \begin{equation*}
 \begin{aligned}
\hspace{-7pt}&\infer[{[I_{\trait}]}]{\trait(\tau, a)}{ \left\{\begin{aligned}
& \sigma_i,  \happens(\action(\alpha_i, a), t_i) \\ & g\big(
  \sigma_i(t)\big) = \sigma, \  g(\alpha_i) = \alpha  \end{aligned}\right\}_{i=1}^n
} \\
\end{aligned}
\end{equation*}
\end{mdframed}
\end{footnotesize}
%%% Local Variables:
%%% mode: latex
%%% TeX-master: "main"
%%% End:

%% file: learningTraits.tex
To address $\mathbf{R}_3$ we need a definition of what it means for an
agent to learn a trait.  We start with a learning agent $l$.  An agent
$e$ is identified as an exemplar by $l$ \emph{iff} the emotion of
admiration is triggered $n$ times or more by $e$ in $l$.  This is
written down in \DCEC\ as follows (note that admiration can be
triggered by different actions):

\begin{footnotesize}
\begin{mdframed}[linecolor=white, frametitle=Exemplar Defninition,
  frametitlebackgroundcolor=gray!10, backgroundcolor=gray!05,
  roundcorner=8pt]
  \begin{equation*}
 \begin{aligned}
\mathit{Exemplar}(e,l) \leftrightarrow \exists^{!n}t. \exists \alpha. \holds(\admires(l,e, \alpha), t)
\end{aligned}
\end{equation*}
\end{mdframed}
\end{footnotesize}

Once $e$ is identified as an exemplar, the learner then identifies one
or more traits of $e$ by observing $e$ over an extended period of
time.  Let $l$ believe that $e$ has a trait $ \tau$; then $l$
incorporates $\tau$ as its own trait:

\begin{footnotesize}
\begin{mdframed}[linecolor=white, frametitle=($\mathbf{R}_3$) Learning a
  Trait,
  frametitlebackgroundcolor=gray!10, backgroundcolor=gray!05,
  roundcorner=8pt]
  \begin{equation*}
 \begin{aligned}
&\mathit{LearnTrait}(l, \tau, t) \leftrightarrow  \exists e \left[\begin{aligned} &\exemplar(e,
l) \land \\ &\believes\Big(l, t, \trait\big(\tau,
e\big)\Big) \end{aligned}\right]\\
& \mathit{LearnTrait}(l, \langle \sigma, \alpha \rangle, t) \rightarrow \big(\sigma \rightarrow  \happens(\action(l,\alpha), t)
\big)
\end{aligned}
\end{equation*}
\end{mdframed}
\end{footnotesize}

\subsubsection{Example}

For instance, if the action type \emph{``being truthful''} is
triggered in situations: \emph{``talking with alice,''},
\emph{``talking with bob''}, \emph{``talking with charlie''}; then the
trait learned is that \emph{``talking with an agent''} situation
should trigger the \emph{``being truthful''} action type.

% A learnt trait is defined below:

% \begin{mdframed}[linecolor=white, frametitle=Learnt Trait,
%   frametitlebackgroundcolor=gray!05, backgroundcolor=gray!02,
%   roundcorner=8pt]
%   A learnt trait is simply a situation $\sigma(t)$ and an action type
%   $\alpha$: $\langle \sigma(t), \alpha \rangle$
% \end{mdframed}

% The learner then simply associates the action type $\alpha$ with the
% generalization of the situations
% $g(\{ \sigma_1 , \sigma_2, \ldots, \sigma_n\})$. That is the agent has
% incorporated this learnt trait:

% $$\Big\langle g\big(\{ \sigma_1, \sigma_2, \ldots, \sigma_n\}\big),
% \alpha \Big\rangle$$

%%% Local Variables:
%%% mode: latex
%%% TeX-master: "main"
%%% End:

%% file: learning.tex
When we look at humans learning virtues by observing others or by
reading from texts or other sources, it is not entirely clear how
models of learning that have been successful in perception and
language processing (e.g.\ the recent successes of deep learning and
statistical learning) can be applied.  Learning in \VE-relevant
situations is from one or few instances or in some cases through
instruction, and such learning may not be readily amenable to models
of learning which require a large number of examples.

The abstract learning method that we will use is \textbf{generalization},
defined previously.  See one simple example immediately below:

\begin{footnotesize}
\begin{mdframed}[linecolor=white, frametitle=Example 1,
  frametitlebackgroundcolor=white, backgroundcolor=white,
  roundcorner=8pt]
  \begin{equation*}
 \begin{aligned}
&\Gamma_1  = \{ \mathit{talkingWith} (\mathit{jack}) \rightarrow \mathit{Honesty} \}\\
&\Gamma_2  = \{ \mathit{talkingWith} (\mathit{jill}) \rightarrow \mathit{Honesty} \}\\
\midrule
\mathbf{generalization } \  &\Gamma  = \{ \forall x. \mathit{talkingWith}(x) \rightarrow  \mathit{Honesty} \}
\end{aligned}
\end{equation*}
\end{mdframed}
\end{footnotesize}

One particularly efficient and well-studied mechanism to realize
generalization is \textbf{anti-unification}, which has been applied
successfully in learning programs from few examples.\footnote{This
  discipline, known as \textbf{inductive programming}, seeks to build
  precise computer programs from examples
  \cite{nienhuys1997foundations}. See \cite{Muggleton2018} for an
  application in generating human comprehensible programs.} In
anti-unification, we are given a set of expressions
$\{f_1, \ldots, f_n\}$ and need to compute an expression $g$ that when
substituted with an appropriate term $\theta_i$ gives us $f_i$.  For
example, if we are given $\mathit{hungry}(\mathit{jack})$ and
$\mathit{hungry}(\mathit{jill})$, the anti-unification of those terms
would be $\mathit{hungry}(\mathit{x})$.

In higher-order anti-unification, we can substitute function symbols
and predicate symbols.  Here $P$ is a higher-order variable.
%% NAVEEEN: Do you want a colon rather than a period to end the
%% previous sentence?  //S

\begin{footnotesize}
  \begin{equation*}
 \begin{aligned}
&\mathbf{Example 2}\\
&\mathit{likes}(\mathit{jill, jack})\\
&\mathit{likes}(\mathit{jill, jim})\\
\midrule
& \mathit{likes}(\mathit{jill, x})
\end{aligned}
\hspace{24pt}
 \begin{aligned}
&\mathbf{Example 3}\\
&\mathit{likes}(\mathit{jill, jack})\\
&\mathit{loves}(\mathit{jill, jim})\\
\midrule
 \ & \mathit{P}(\mathit{jill, x}) 
\end{aligned}
\end{equation*}
\end{footnotesize}

%%% Local Variables:
%%% mode: latex
%%% TeX-master: "main"
%%% End:

%% file: topdefs.tex
With the formal machinery in place we finally present formalizations
that answer $\mathbf{Q}_1$ and $\mathbf{Q}_2$ posed at the outset.  An
$n$-virtuous person or agent $s$ is an agent that is considered as an
exemplar by $n$ agents:
%% NAVEEEN: I tend to think that the period ending the previous
%% sentence should be a colon.  //S

\begin{footnotesize}
\begin{mdframed}[linecolor=white, frametitle=($\mathbf{Q}_1$) Virtuous Person,
  frametitlebackgroundcolor=gray!10, backgroundcolor=gray!05,
  roundcorner=8pt]
    \begin{equation*}
\mathbf{V}_n(s) \leftrightarrow \exists^{\geq n} a:\exemplar(s, a)
  \begin{aligned}
 \end{aligned}
\end{equation*}
\end{mdframed}
\end{footnotesize}

\noindent An $n$-virtue is a trait possesed by at least $n$ virtuous
agents: 
%% NAVEEEN: I tend to think that the period ending the previous
%% sentence should be a colon.  //S
\begin{footnotesize}
\begin{mdframed}[linecolor=white, frametitle=($\mathbf{Q}_2$) Virtue,
  frametitlebackgroundcolor=gray!10, backgroundcolor=gray!05,
  roundcorner=8pt]
    \begin{equation*}
\mathbf{G}_n(\tau) \leftrightarrow \exists^{\geq n} a: \trait(\tau, a)
  \begin{aligned}
 \end{aligned}
\end{equation*}
\end{mdframed}
\end{footnotesize}

%%% Local Variables:
%%% mode: latex
%%% TeX-master: "main"
%%% End:

%% file: example.tex
We have extended \emph{ShadowProver}, a quantified modal logic prover
for \DCEC\ used in \cite{nsg_sb_dde_2017} to handle the new inference
schemata and definitional axioms given above.  We now show a small
simulation in which an agent learns a trait and uses that trait to
perform an action. Assume that we have a marketplace where things that
are either old or new can be bought and sold.  A seller can either
honestly state the condition of an item
$\{\mathit{new}, \mathit{old}\}$ or falsely report the state of the
item. Agent $a$ has two items $x$ and $y$. $x$ is new and $y$ is
old. $a$ is asked about the state of the items, and $a$ responds
accurately. We have an agent $d$ that observes agent $a$ correctly
report the state of the items. $d$ also has beliefs about $a$'s state
of mind. We also have that the agent $d$ considers $a$ to be an
exemplar. When all this information is fed into the prover along with
the definitions above, $d$ learns a trait representing a form of
honesty, shown below:
 
% \begin{footnotesize}
%     \begin{equation*}
%       \begin{aligned}
%          & \textbf{Situation 1}\\
%           &\holds(old,t) \\
%            & \happens(utter(broken old), t)\\
%       \end{aligned}
%       \begin{aligned}
%          & \textbf{Situation 2}
%         & \holds(new,t) \\
%         &  \happens(utter(new), t)
%       \end{aligned}
%     \end{equation*}
%   \end{footnotesize}

\begin{footnotesize}
\begin{mdframed}[linecolor=white,  frametitlebackgroundcolor=white, backgroundcolor=white,
  roundcorner=8pt]
    \begin{equation*}
 \Bigg\langle \begin{aligned}
 &\believes(d, t, \holds(x,t) \land \nu( \mathit{utter}
 (x), t) > 0),  \\ &\mathit{utter}(x) 
 \end{aligned}\Bigg\rangle
\end{equation*}
\end{mdframed}
\end{footnotesize}

\begin{figure}[h!]
 \centering
 {
  \shadowbox{\includegraphics[width=\linewidth]{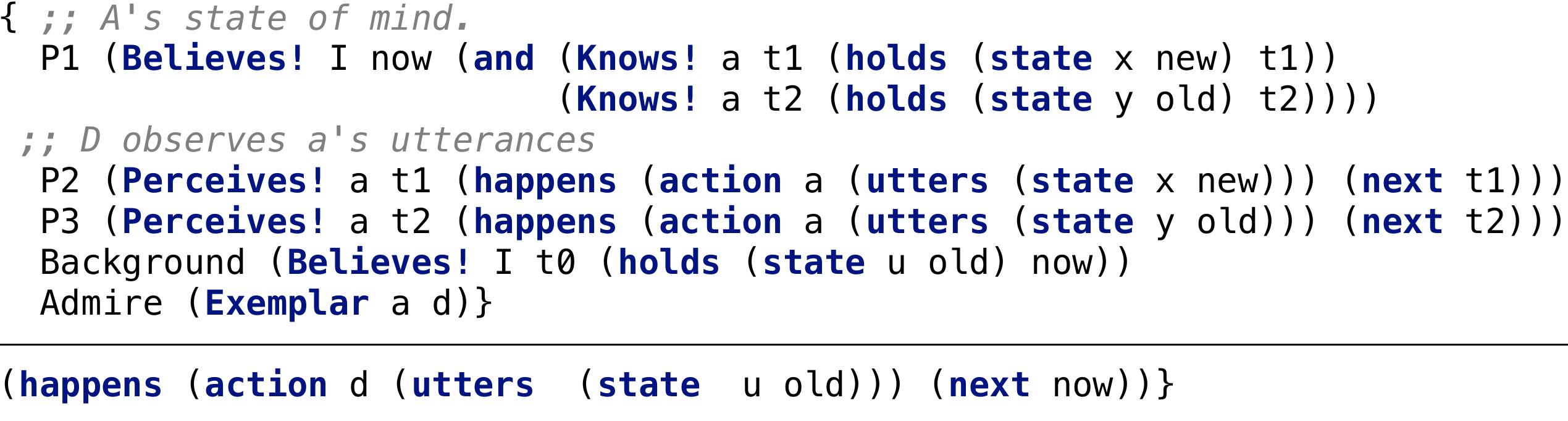}}}
\caption{Simulation Input and Output Formulae}
 \label{fig:ccOverview}
\end{figure}

 When $d$ is
queried about the state of an item $u$, $d$ responds accurately (input
and output shown in Figure ~\ref{fig:ccOverview}). The prover responds
with the required output in $3.6$ seconds.\footnote{A notebook for
  running this simulation can be found here: \url{anonymized}.}

 %%++++++++++++++++++++++++++++++++++++++++++++++++++++++++++++++++++++
%%++++++++++++++++++++++++++++++++++++++++++++++++++++++++++++++++++++
%% NAVEEN: `learnt' rather than `learned' would I think by British.  I
%% changed others, but not this one.  Can you change?  Thx.  //S
%The learnt trait is then given below. The trait says that one should
%always correctly utter the state of the item.

% \begin{footnotesize}
% \begin{mdframed}[linecolor=white,  frametitlebackgroundcolor=white, backgroundcolor=white,
%   roundcorner=8pt]
%     \begin{equation*}
%  \begin{aligned}
%  \Big\langle \holds(x,t),  \happens(utter(x), t) \Big\rangle
%  \end{aligned}
% \end{equation*}
% \end{mdframed}
% \end{footnotesize}

%%% Local Variables:
%%% mode: latex
%%% TeX-master: "main"
%%% End: